\documentclass[10pt,twocolumn,letterpaper]{article}

\usepackage{iccv}
\usepackage{times}
\usepackage{epsfig}
\usepackage{graphicx}
\usepackage{amsmath}
\usepackage{amssymb}
\usepackage{xcolor}
\usepackage{authblk}
\makeatletter
\renewcommand\AB@affilsepx{ \;\; \protect\Affilfont}
\makeatother

\usepackage[pagebackref=true,breaklinks=true,letterpaper=true,colorlinks,bookmarks=false]{hyperref}

\iccvfinalcopy 

\def\iccvPaperID{4130} 

\newcommand{\baenet}{\textsc{Bae-Net}\xspace}

\newcommand{\zq}[1]{{\color{black}#1}}
\newcommand{\rz}[1]{{\color{black}#1}}

\ificcvfinal\fi
\begin{document}

\title{\textbf\baenet: Branched Autoencoder for Shape Co-Segmentation}

\author[1]{Zhiqin Chen}
\author[1]{Kangxue Yin}
\author[2]{Matthew Fisher}
\author[2,3]{Siddhartha Chaudhuri}
\author[1]{Hao Zhang}
\affil[1]{Simon Fraser University}
\affil[2]{Adobe Research}
\affil[3]{IIT Bombay}

\maketitle
\thispagestyle{empty}

\begin{abstract}
We treat shape co-segmentation as a {\em representation learning\/} problem and
introduce \baenet, a {\em branched autoencoder\/} network, for the task. The unsupervised \baenet
is trained with a collection of {\em un-segmented\/} shapes, using a {\em shape\/} 
reconstruction loss, without any ground-truth labels. Specifically, the network takes an input shape and encodes
it using a convolutional neural network, whereas the decoder concatenates the resulting
feature code with a point coordinate and outputs a value indicating whether the point is
inside/outside the shape. Importantly, the decoder is branched: each branch learns a compact
representation for one commonly recurring part of the shape collection, e.g., airplane wings. By complementing
the shape reconstruction loss with a label loss, \baenet is easily tuned for {\em one-shot\/}
learning. We show unsupervised, weakly supervised, and one-shot learning results by \baenet, demonstrating 
that using only a couple of exemplars, our network can generally outperform state-of-the-art 
supervised methods trained on hundreds of segmented shapes.
Code is available at \href{https://github.com/czq142857/BAE-NET}{https://github.com/czq142857/BAE-NET}.
\end{abstract}

\section{Introduction}
\label{sec:intro}

Co-segmentation takes a collection of data sharing some common characteristic and
produces a {\em consistent\/} segmentation of each data item. Specific to
shape co-segmentation, the common characteristic of the input collection is typically tied
to the common category that the shapes belong to, e.g., they are all lamps or chairs. The
significance of the problem is attributed to the consistency requirement, since a shape co-segmentation not
only reveals the structure of each shape but also a structural {\em correspondence\/} across
the entire shape set, enabling a variety of applications including attribute transfer
and mix-n-match modeling.

In recent years, many deep neural networks have been developed for
segmentation~\cite{SegNet, huang2018recurrent, kalogerakis20173d, long2015, pointnet++, yi2017syncspeccnn}.
Most methods to date formulate segmentation as a supervised classification problem and are applicable to segmenting a single
input. Representative approaches include SegNet~\cite{SegNet} for images and PointNet~\cite{pointnet} for shapes,
where the networks are trained by supervision with ground-truth segmentations to map pixel or point features to segment labels.

Co-segmentation seeks a structural understanding or explanation of an entire set. If one were to abide by
Occam's razor, then the best explanation would be the {\em simplest\/} one. This motivates us to
treat co-segmentation as a {\em representation learning\/} problem, with the added bonus that
such learning may be unsupervised without any ground-truth labels. Given
the strong belief that object recognition by humans is part-based~\cite{hebb1949,hoffman1984}, the simplest explanation
for a collection of shapes belonging to the same category would be a combination of universal parts for
that category, e.g., chair backs or airplane wings. Hence, an unsupervised shape
co-segmentation would amount to finding the {\em simplest part representations\/} for a shape collection.
Our choice for the representation learning module is a variant of {\em autoencoder\/}.

\begin{figure}[t!]
\begin{center}
\includegraphics[width=1.0\linewidth]{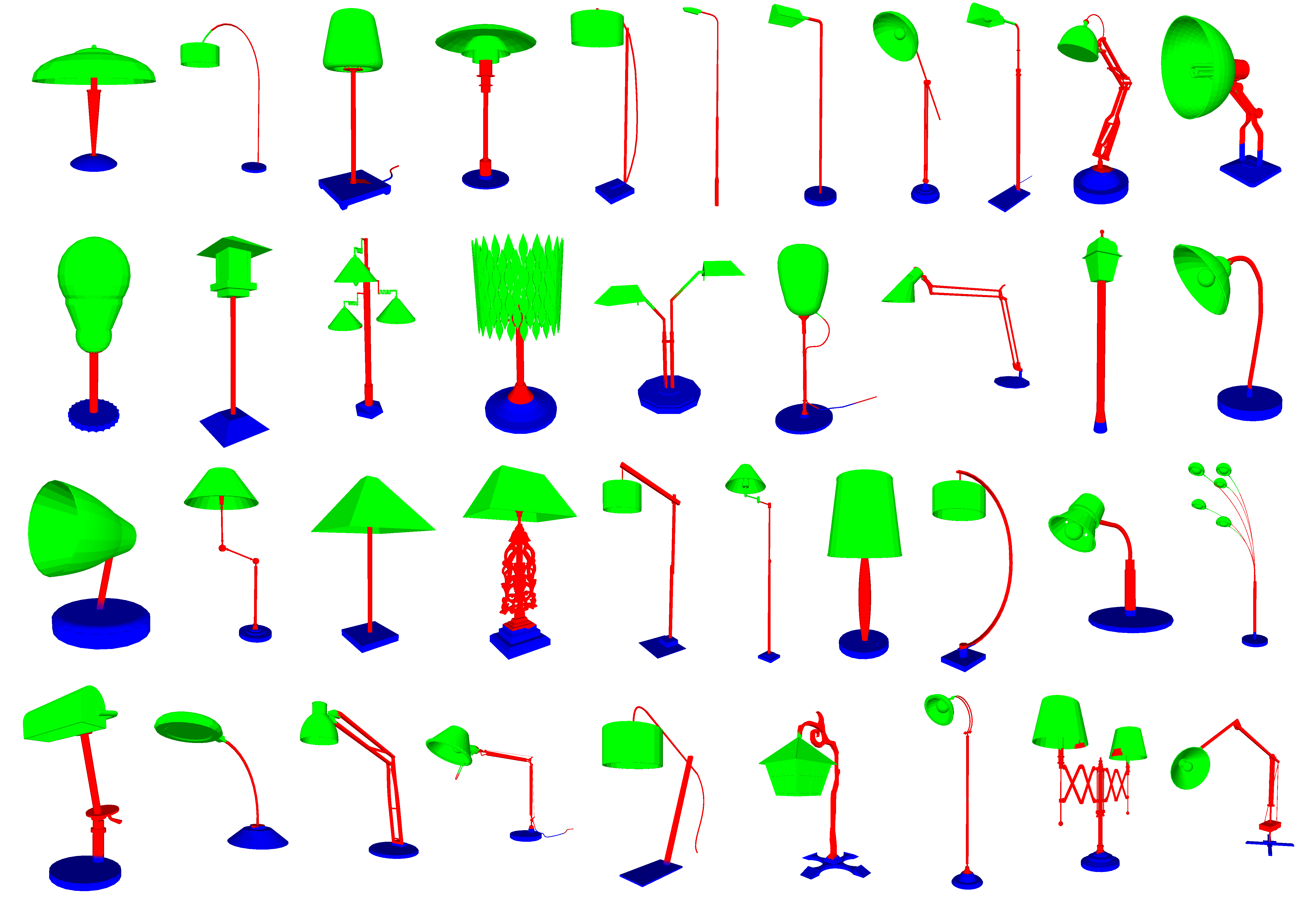}
\end{center}
   \caption{Unsupervised co-segmentation by \baenet on the lamp category from the ShapeNet part dataset \cite{yi2016scalable}. Each color denotes a part labeled by a specific branch of our network.}
\label{fig:teaser}
\vspace{-3mm}
\end{figure}

In principle, autoencoders learn compact representations of a set of data via dimensionality reduction while
minimizing a self-reconstruction loss. To learn shape parts, we introduce a {\em branched\/}
version of autoencoders, where each branch is tasked to learn a simple representation for one universal part
of the input shape collection. In the absence of any ground-truth segmentation labels, our {\em branched
autoencoder\/}, or \baenet for short, is trained to minimize a {\em shape\/} (rather than label) reconstruction
loss, where shapes are represented using implicit fields~\cite{IMGAN}. Specifically, the \baenet decoder takes
as input a concatenation between a point coordinate and an encoding of the input shape (from the \baenet
encoder) and outputs a value which indicates whether the point is inside/outside the shape.

\begin{figure}[t!]
\begin{center}
\includegraphics[width=1.0\linewidth]{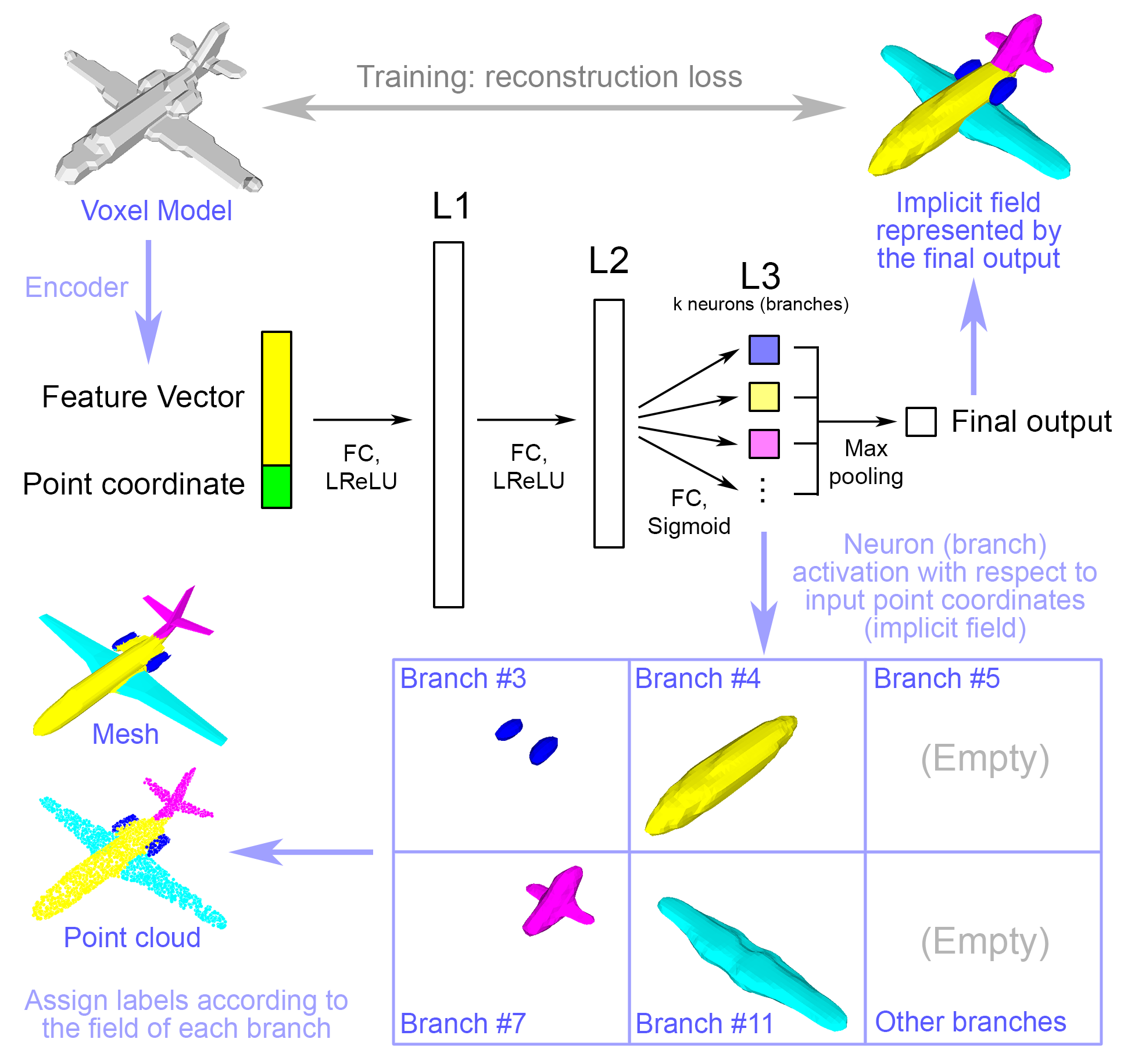}
\end{center}
   \caption{Network architecture of the \baenet decoder; encoder is a CNN. L3 is the branch output layer  (one neuron per branch) that gives the implicit field for each branch. The final output layer groups the branch outputs, via max pooling, to form the final implicit field. Each branch either represents a shape part or simply outputs nothing, if all the parts are represented by other branches. The max pooling operator allows part overlap, giving \baenet the freedom to represent each part in the most natural or simplest way.  All the colors for the parts are for visualization only.}
\label{fig:decoderStructure}
\vspace{-3mm}
\end{figure}

\begin{figure}[t!]
\begin{center}
\includegraphics[width=0.9\linewidth]{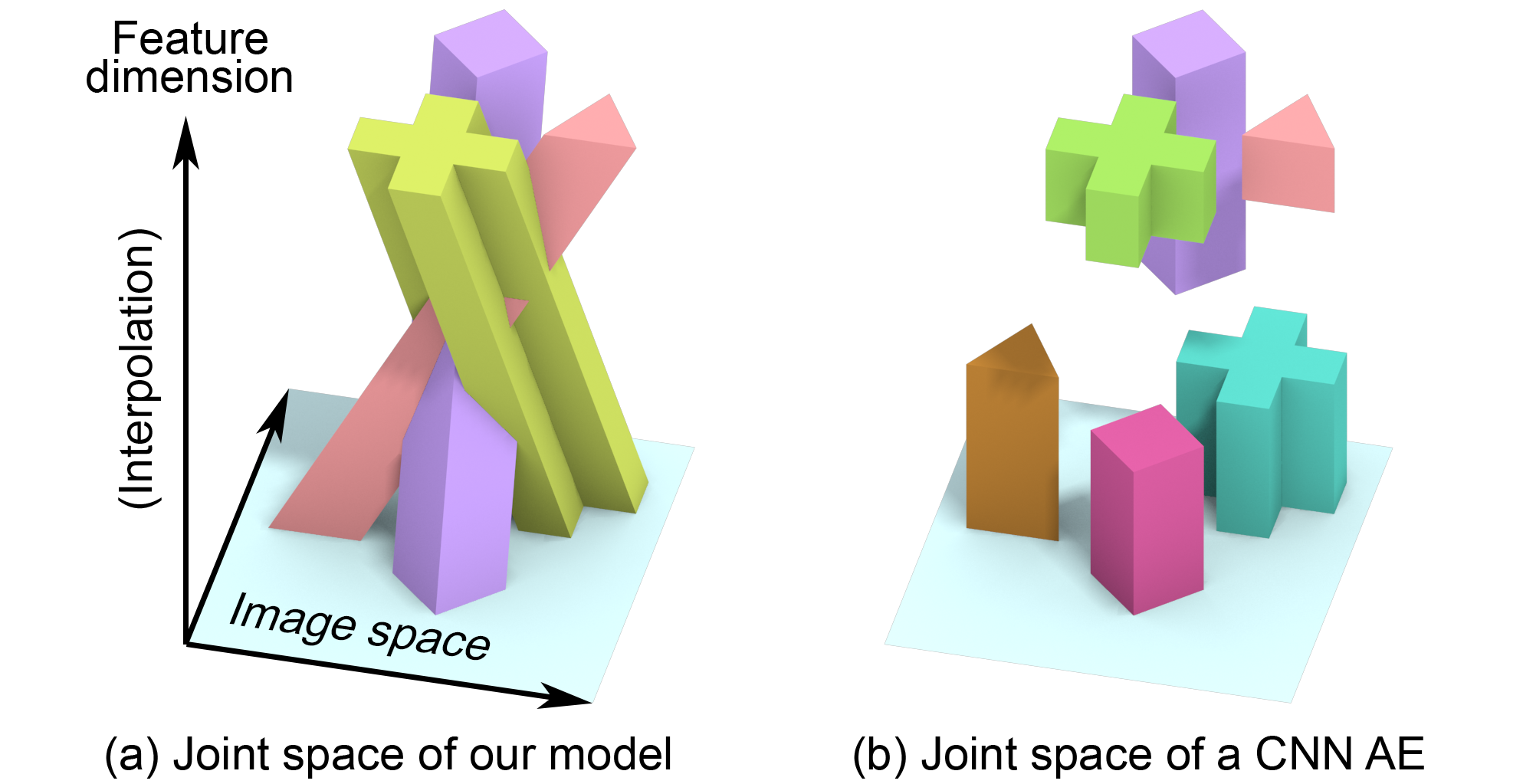}
\end{center}
   \caption{ A toy example where the input set comprises 2D images of three randomly placed patterns. In our network, when the feature dimension is set to be $16$, learning the features amounts to ``sorting'' the training images in the joint (2+16)-D space, so that the hyperdimensional shape is simple and easy to represent. As shown on the left, where only one feature dimension is drawn, \baenet learns a simple three-part representation exhibiting continuity and linearity in the feature dimension, since we treat image and feature dimensions equally in the network. The CNN model on the right represents the shapes as discrete image blocks without continuity. Section~\ref{subsec:shape_extraction} shows results on the full image. }
\label{fig:element3d}
\vspace{-3mm}
\end{figure}

The \baenet architecture is shown in Figure~\ref{fig:decoderStructure}, where the encoder employs a
traditional convolutional neural network (CNN). The encoder-decoder combination of \baenet is trained with all
shapes in the input collection using a (shape) reconstruction loss. Appending the point coordinate to the decoder input
is critical since it adds {\em spatial awareness\/} to the reconstruction process, which is often lost
in the convolutional features from the CNN encoder. Each neuron in the third layer (L3) is trained to learn the inside/outside status of the point relative to one shape part. The parts are learned in a {\em joint\/} space of shape features and point
locations. In Section~\ref{sec:results}, we show that the limited neurons in our autoencoder for representation learning, and the linearities modeled by the neurons, allow \baenet to learn compact part representations in the decoder
branches in L3.
Figure~\ref{fig:element3d} shows a toy example to illustrate the joint space and contrasts the part representations
that are learned by our \baenet versus a classic CNN autoencoder. 

\baenet has a simple architecture and as such, it can be easily adapted to perform {\em one-shot\/}
learning, where only one or few exemplar segmentations are provided. In this case, the number of
branches is set according to the exemplar(s) and the shape reconstruction loss is {\em complemented\/} by
a label reconstruction loss; see Section~\ref{subsec:loss}.

We demonstrate unsupervised, weakly supervised, and one-shot learning results for shape co-segmentation on the ShapeNet~\cite{chang2015shapenet}, ShapeNet part~\cite{yi2016scalable}, and Tags2Parts~\cite{tags2parts} datasets, comparing \baenet with existing supervised and weakly supervised methods. The co-segmentation results from \baenet are consistent without explicitly enforcing any consistency loss in the network. Using only one (resp. two or three) segmented exemplars, the one-shot learning version of \baenet outperforms state-of-the-art supervised segmentation methods, including PointNet++ and PointCNN, when the latter are trained on 10\% (resp. 20\% or 30\%), i.e., hundreds, of the segmented shapes.

\section{Related work}
\label{sec:related}

Our work is most closely related to prior research on unsupervised, weakly supervised, and semi-supervised co-segmentation of 3D shapes. Many of these methods, especially those relying on statistical learning, are inspired by related research on 2D image segmentation. 

\vspace{5pt}

\noindent
{\bf Image co-segmentation without strong supervision.}~
One may view unsupervised image co-segmentation as pixel-level clustering, guided by color similarity within single images as well as consistency across image collections. Existing approaches employ graph cuts~\cite{rother2006coseg}, discriminative clustering~\cite{joulin2010cluster,joulin2012multiclass}, correspondences~\cite{rubinstein2013unsup,rubio2012region}, cooperative games~\cite{lin2014game}, and deep neural networks~\cite{hsu2018attention,wang2017icassp,yu2018sp}.

An alternative research theme utilizes weak cues, such as image-level labels. In this {\em weakly supervised} setup, the goal is to find image regions that strongly correlate with each label, using either traditional statistical models~\cite{cinbis2017multifold,song2014localize,wang2014latent} or newer models based on deep neural networks~\cite{ahn2018affinity,durand2017wildcat,ge2018fusion,huang2018deepseed,kolesnikov2016three,oquab2015weakcnn,pathak2015constrained,pathak2015fcmil,pinheiro2015weakseg,saleh2016fgbg,wang2018weakseg,zhang2018acl}. Other forms of weak supervision include bounding boxes~\cite{dai2015boxsup,hu2018segevery,khoreva2017simple,papandreou2015weaksemi,rajchl2017deepcut,zhao2018pseudomask} and textual captions~\cite{berg2004namesfaces,everingham2009naming}. {\em Semi-supervised} methods assume that full supervision is available for only a few images, while the others are unsupervised: such approaches have been applied to image co-segmentation, e.g.~\cite{ma2013transduction,wang2013semisup}.

In contrast to all the above methods, we develop an unsupervised co-segmentation approach for geometric {\em shapes\/}, without colors. Thus, we concentrate on efficient modeling of {\em spatial} variance. We employ a novel encode-and-reconstruct scheme, where each branch of a deep network learns to localize instances of a part across multiple examples in order to compactly represent them, and re-assemble them into the original shape. Our method easily adapts to weakly- and semi-supervised scenarios.

Our method is critically dependent on the choice of network architecture. The relatively shallow fully-connected stack is high-capacity enough to model non-trivial parts, but shallow enough that the neurons are forced to learn a compact, efficient representation of the shape space, in terms of recurring parts carved out by successive simple units. Thus, the geometric prior is inherent in the architecture itself, \zq{similar in spirit to Deep Image Prior~\cite{DeepImPr} and Deep Geometric Prior for Surface Reconstruction~\cite{williams2019deep}}.

\vspace{5pt}

\noindent
{\bf 3D segmentation without strong supervision.}~
Building on substantial prior work on single-shape mesh segmentation based on geometric cues~\cite{shamir2008segmentation}, the pioneering work of Golovinskiy and Funkhouser~\cite{golovinskiy2009consistent} explored consistent co-segmentation of 3D shapes by constructing a graph connecting not just adjacent polygonal faces in the same mesh, but also corresponding faces across different meshes. A normalized cut of this graph yields a joint segmentation. Subsequently, several papers developed alternative unsupervised clustering strategies for mesh faces, given a handcrafted similarity metric induced by a feature embedding or a graph~\cite{hu2012co,huang2011joint,meng2013iterative,sidi2011unsupervised,xu2010style}. Shu et al.~\cite{shu2016unsup} modified this setup by first transforming handcrafted local features with a stacked autoencoder before applying an (independently learned) Gaussian mixture model and per-shape graph cuts. In contrast, our method is an end-to-end differentiable pipeline that requires no manual feature specification or large-scale graph optimization.

Tulsiani et al.~\cite{tulsiani2017learning} proposed an unsupervised method to approximate a 3D shape with a small set of primitive parts (cuboids), inducing a segmentation of the underlying mesh. Their approach has similarities to ours -- they predict part cuboids with branches from an encoder network, and impose a reconstruction loss to make the cuboid assembly resemble the original shape. However, the critical difference is the restriction to cuboidal boxes: they cannot accommodate complex, non-convex part geometries such as the rings in Figure~\ref{fig:2D_compare} and the groups of disjoint lamp parts in Figure~\ref{fig:teaser}, for which a nonlinear stack of neurons is a much more effective indicator function.

In parallel, approaches were developed for weakly supervised~\cite{tags2parts,shilane2007distinctive,zhu2019cosegnet} and semi-supervised~\cite{Huang:2015:deeplearningsurfaces,kim13templates,wang2012active} shape segmentation. Shapes formed by articulation of a template shape can be jointly co-segmented~\cite{anguelov2004articulated,yi2018articulated}. Our method does not depend on any such supervision or base template, although it can optionally benefit from one or two annotated examples to separate strongly correlated part pairs.





\section{\baenet: architecture, loss, and training}
\label{sec:method}


The architecture of \baenet draws inspiration from IM-NET, the implicit decoder recently introduced by Chen and Zhang~\cite{IMGAN}; \rz{also see concurrent works~\cite{deepSDF,OccNet} proposing similar ideas.} IM-NET learns an implicit field by means of a binary classifier, which is similar to \baenet. The main difference, as shown in Figure~\ref{fig:decoderStructure}, is that \baenet is designed to segment a shape into different parts by reconstructing the parts in different branches of the network.

Similar to~\cite{IMGAN}, we use a traditional convolutional neural network as the encoder to produce the feature code for a given shape. We also adopt a three-layer fully connected neural network as our decoder network. The network takes a joint vector of point coordinates and feature code as input, and outputs a value in each output branch that indicates whether the input point is inside a part (1) or not (0). Finally, we use a max pooling operator to merge parts together and obtain the entire shape, which allows our segmented parts to overlap. We use ``L1'', ``L2'' and ``L3'' to represent the first, second, and third layer, respectively. The different network design choices will be discussed in Section~\ref{sec:results}.

\subsection{Network losses for various learning scenarios}
\label{subsec:loss}

\paragraph{Unsupervised.}
For each input point coordinate, our network outputs a value that indicates the likelihood that the given point is inside the shape. We train our network with sampled points in the 3D space surrounding the input shape and the inside-outside status of the sampled points. After sampling points for input shapes using the same method as ~\cite{IMGAN}, we can train our autoencoder with a mean square loss:
\begin{align}
\mathcal{L}_\mathrm{unsup} (\mathbb{P} (\mathrm{S})) = \mathbb{E}_{\mathrm{S} \sim \mathbb{P} (\mathrm{S})} \mathbb{E}_{p \sim \mathbb{P} (p|\mathrm{S})} (f(p) - \mathrm{F}(p))^2
\end{align}
where $\mathbb{P} (\mathrm{S})$ is the distribution of training shapes, $\mathbb{P} (p|\mathrm{S})$ is the distribution of sampled points given shape $\mathrm{S}$, $f(p)$ is the output value of our decoder for input point $p$, and $\mathrm{F}(p)$ is the ground truth inside-outside status for point $p$. This loss function allows us to reconstruct the target shape in the output layer. The segmented parts will be expressed in the branches of L3, since the final output is taken as the maximum value over the fields represented by those branches.

\vspace{-10pt}

\paragraph{Supervised.}
If we have examples with ground truth part labels, we can also train \baenet in a supervised way. Denote $\mathrm{F}_m(p)$ as the ground truth status for point $p$, and $f_m(p)$ as the output value of the $m$-th branch in L3. For a network with $k$ branches in L3, the supervised loss is:
\begin{align*}
\mathcal{L}_\mathrm{sup*} (\mathbb{P} (\mathrm{S})) = \mathbb{E}_{\mathrm{S} \sim \mathbb{P} (\mathrm{S})} \mathbb{E}_{p \sim \mathbb{P} (p|\mathrm{S})} \frac{1}{k}\sum_{i=1}^{k}(f_i(p) - \mathrm{F}_i(p))^2
\end{align*}
In datasets such as the ShapeNet part dataset \cite{yi2016scalable}, shapes are represented by point clouds sampled from their surfaces. In such datasets, the inside-outside status of a point can be ambiguous. However, since our sampled points are from voxel grids and the reconstructed shapes are thicker than the original, we can assume all points in the ShapeNet part dataset are inside our reconstructed shapes. We use both our sampled points from voxel grids and the point clouds in the ShapeNet part dataset, by modifying the loss function:
\begin{align}
\begin{aligned}
\mathcal{L}_\mathrm{sup} (\mathbb{P} (\mathrm{S})) & = \mathbb{E}_{\mathrm{S} \sim \mathbb{P} (\mathrm{S})} [ \mathbb{E}_{p \sim \mathbb{P} (p|\mathrm{S})} (f(p) - \mathrm{F}(p))^2 \\
 & + \alpha \mathbb{E}_{q \sim \mathbb{P} (q|\mathrm{S})} \frac{1}{k}\sum_{i=1}^{k}(f_i(q) - \mathrm{F}_i(q))^2 ]
\end{aligned}
\end{align}
where $\mathbb{P} (p|\mathrm{S})$ is the distribution for our sampled points from voxel grids, and $\mathbb{P} (q|\mathrm{S})$ is the distribution of points in the ShapeNet part dataset. We set $\alpha$ to 1 in our experiments.

\vspace{-10pt}

\paragraph{One-shot learning.}
Our network also supports one-shot training, where we feed the network 1, 2, or 3, $\ldots$, shapes with ground truth part labels, and other shapes without part labels. To enable one-shot training, we have a joint loss:
\begin{align}
\mathcal{L}_\mathrm{joint} = \mathcal{L}_\mathrm{unsup} (\mathbb{P} (\mathrm{S})) + \beta \mathcal{L}_\mathrm{sup} (\mathbb{P} (\mathrm{S'}))
\end{align}
where $\mathbb{P} (\mathrm{S})$ is the distribution of all shapes, and $\mathbb{P} (\mathrm{S'})$ is the distribution of the few given shapes with part labels. We do not explicitly use this loss function or set $\beta$. Instead, we train the network using the unsupervised and supervised losses alternately. We do one supervised training iteration after every 4 unsupervised training iterations.

Additionally, we add a very small $L1$ regularization term for the parameters of L3 to prevent unnecessary overlap, e.g., when the part output by one branch contains the part output by another branch.

\subsection{Point label assignment}

After training, we get an approximate implicit field for the input shape. To label a given point of an input shape, we simply feed the point into the network together with the code encoding the feature of the input shape, and label the point by looking at which branch in L3 gives the highest value. If the training has exemplar shapes as guidance, each branch will be assigned a label automatically with respect to the exemplars. If the training is unsupervised, we need to look at the branch outputs and give a label for each branch by hand. For example in Figure~\ref{fig:decoderStructure}, we can label branch \#3 as ``jet engine'', and each point having the highest value in this branch will be labeled as ``jet engine''. To label a mesh model, we first subdivide the surfaces to obtain fine-grained triangles, and assign a label for each triangle. To label a triangle, we feed its three vertices into the network and sum their output values in each branch, and assign the label whose associated branch gives the highest value.

\subsection{Training details}
\label{subsec:train_details}

In what follows, we denote the decoder structure by the width (number of neurons) of each layer as \mbox{\{ L1-L2-L3 \}}. The encoders for all tasks are standard CNN encoders.

In the 2D shape extraction experiment, we set the feature vector to be 16-D, since the goal of this experiment is to explain why and how the network works and the shapes are easy to represent. We use the same width for all hidden layers since it is easier for us to compare models with different depths. For \mbox{$64 \times 64$} images, the decoder is \mbox{\{ 256-4 \}} for 2-layer model, \mbox{\{ 256-256-4 \}} for 3-layer model, and \mbox{\{ 256-256-256-4 \}} for 4-layer model. For \mbox{$128 \times 128$} images, we use $512$ as the width instead of $256$.

In all other experiments, our encoder takes $64^3$ voxels as input, and produces a $128$-D feature code. We sample $8192$ points from each shape's $32^3$ voxel model, and use these point-value pairs to compute the unsupervised loss $\mathcal{L}_\mathrm{unsup}$.

For unsupervised tasks, we set the decoder network to \mbox{\{ 3072-384-12 \}} and train 200,000 iterations with mini-batches of size 1, which takes $\sim$2 hours per category.
For one-shot tasks, we use a \mbox{\{ 1024-256-$n$ \}} decoder, where $n$ is the number of ground truth parts in exemplar shapes. The decoder is lighter, hence we finish training in a shorter time. For each category, we train 200,000 iterations: on all 15 categories this takes less than 20 hours total on a single NVIDIA GTX 1080Ti GPU. We also find that doing 3,000-4,000 iterations of supervised pretraining before alternating it with unsupervised training improves the results.

\section{Experiments and results}
\label{sec:results}

In this section, we first discuss different architecture design choices and offer insights into how our network works. Further, we show qualitative and quantitative segmentation
results in various settings with \baenet and compare them to those from state-of-the-art methods.


\subsection{Network design choices and insights}
\label{subsec:shape_extraction}

\begin{figure}[t!]
\begin{center}
\includegraphics[width=1.0\linewidth]{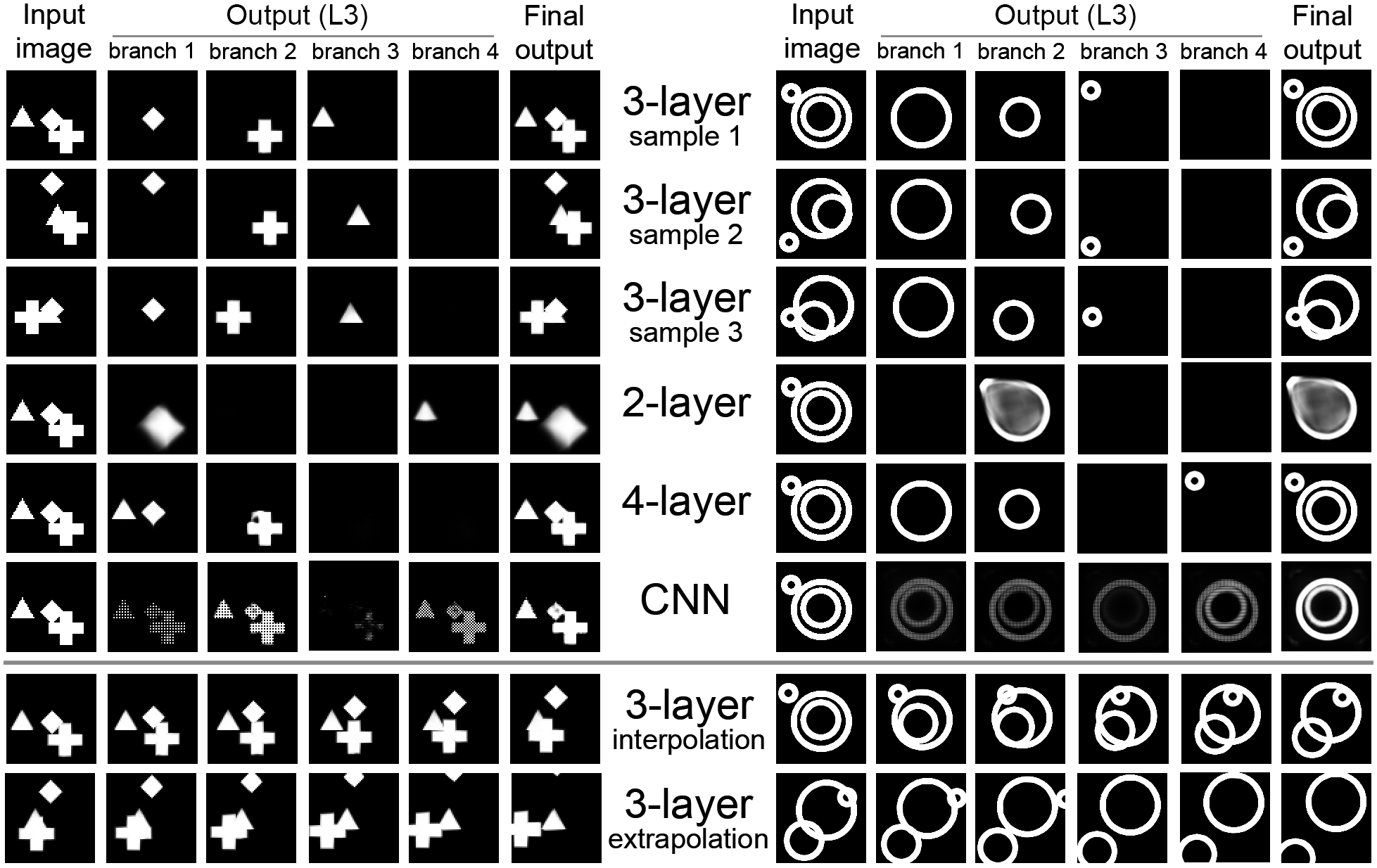}
\end{center}
   \caption{Independent shape extraction results of different models. The first three rows show the segmentation results of the 3-layer model. The next three rows show the segmentation results of other models for comparison. The last row shows the extrapolation results continuing its previous row. Note that no shape patterns go beyond the boundary in our synthesized training dataset, thus we can be certain that some shapes in the last row are completely new.}
\label{fig:2D_compare}
\end{figure}

\begin{figure}[t!]
\begin{center}
\includegraphics[width=\linewidth]{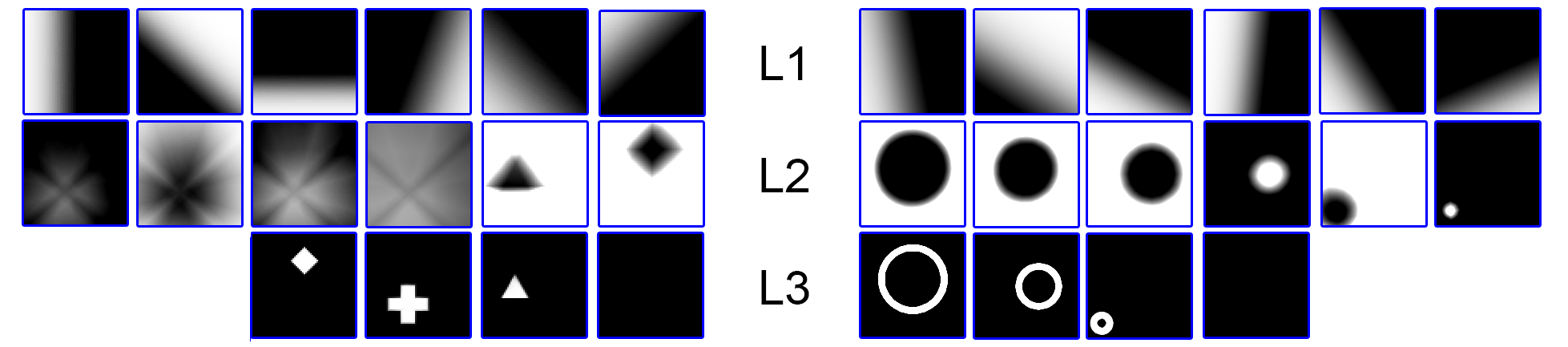}
\end{center}
   \caption{Visualization of neurons in the first, second and third layer of our 3-layer network. Since L1 and L2 have hundreds of neurons, we only select a few representative ones to show here. More visualizations can be found in the supplementary material.}
\label{fig:2D_vis}
\vspace{-3mm}
\end{figure}

We first explain our network design choices in detail, as illuminated by two synthetic 2D datasets: ``elements'' and ``triple rings''. ``Elements'' is synthesized by putting three different shape patterns over $64 \times 64$ images, where the cross is placed randomly on the image, the triangle is placed randomly on a horizontal line in the middle of the image, and the rhombus is placed randomly on a vertical line in the middle of the image. ``Triple rings'' is synthesized by placing three rings of different sizes randomly over $128 \times 128$ images. See Figure~\ref{fig:2D_compare} for some examples.

First, we train \baenet with 4 branches on the two datasets; see some results in Figure~\ref{fig:2D_compare}. Our network successfully separated the shape patterns, even when two patterns overlap. Further, each of the output branches only outputs one specific shape pattern, thus we also obtain a shape correspondence from the co-segmentation process.

We visualize the neuron activations in Figure~\ref{fig:2D_vis}. In L1, the point coordinates and the shape feature code have gone through a linear transform and a leaky ReLU activation, therefore the activation maps in L1 are linear ``space dividers'' with gradients. In L2, each neuron linearly combines the fields in L1 to form basic shapes. The combined shapes are mostly convex: although non-convex shapes can be formed, they will need more resources (L1 neurons) than simpler shapes. This is because L2 neurons calculate a weighted sum of the values in L1 neurons, not MIN, MAX, or a logical operation, thus each L1 neuron brings a global, rather than local, change in L2. L2 represents higher level shapes than L1, therefore we can expect L1 to have many more neurons than L2, and we incorporate this idea in our network design for shape segmentation, to significantly shorten training time. The L3 neurons further combine the shapes in L2 to form output parts in our network, and our final output combines all L3 outputs via max pooling.

These observations and insights offer some explanation as to why our network tends to output segmented, corresponding parts in each branch. For a single shape, the network has limited representation ability in L3, therefore it prefers to construct simple parts in each branch, and let our max pooling layer combine them together. This allows better reconstruction quality than reconstructing the whole shape in just one branch. With an appropriate optimizer to minimize the reconstruction error, we can obtain well-segmented parts in the output branches.

For part correspondence, we need to also consider the input shape feature codes. As shown in Figure~\ref{fig:decoderStructure}, our network treats the feature code and point coordinates equally. This allows us to consider the whole decoder as a hyperdimensional implicit field, in a joint space made by both image dimensions (input point coordinates) and shape feature dimensions (shape feature code). In Figure~\ref{fig:element3d}, we visualize a 3D slice of this implicit field, with two image space dimensions and one feature dimension. Our network is trying to find the best way to represent all shapes in the training set, and the easiest way is to arrange the training shapes so that the hyperdimensional shape is continuous in the feature dimension, as shown in the figure. This encourages the network to learn to represent the final complex hyperdimensional shape as a composition of a few simple hyperdimensional shapes. In Figure~\ref{fig:2D_compare}, we show how our trained network can accomplish smooth interpolation and extrapolation of shapes. Our network is able to simultaneously accomplish segmentation and correspondence.

We compare the segmentation results of our current 3-layer model with a 2-layer model, a 4-layer model, and a CNN model in Figure~\ref{fig:2D_compare}; detailed network parameters are in supplementary material. The 2-layer model has a hard time reconstructing the rings, since L2 is better at representing convex shapes. The 4-layer model can separate parts, but since most shapes can already be represented in L3, the extra layer does not necessarily output separated parts. One can easily construct an L4 layer on top of our 3-layer model to output the whole shape in one branch while leaving the other branches blank. The CNN model is not sensitive to parts and outputs basically everything or nothing in each branch, since there is no bias towards sparsity or segmentation. Overall, the 3-layer network is the best choice for independent shape extraction, making it a suitable candidate for unsupervised and weakly supervised shape segmentation.

\subsection{Evaluation of unsupervised learning} 
\label{subsec:unsupervised_segmentation}

\begin{figure}[t!]
\begin{center}
\includegraphics[width=1.0\linewidth]{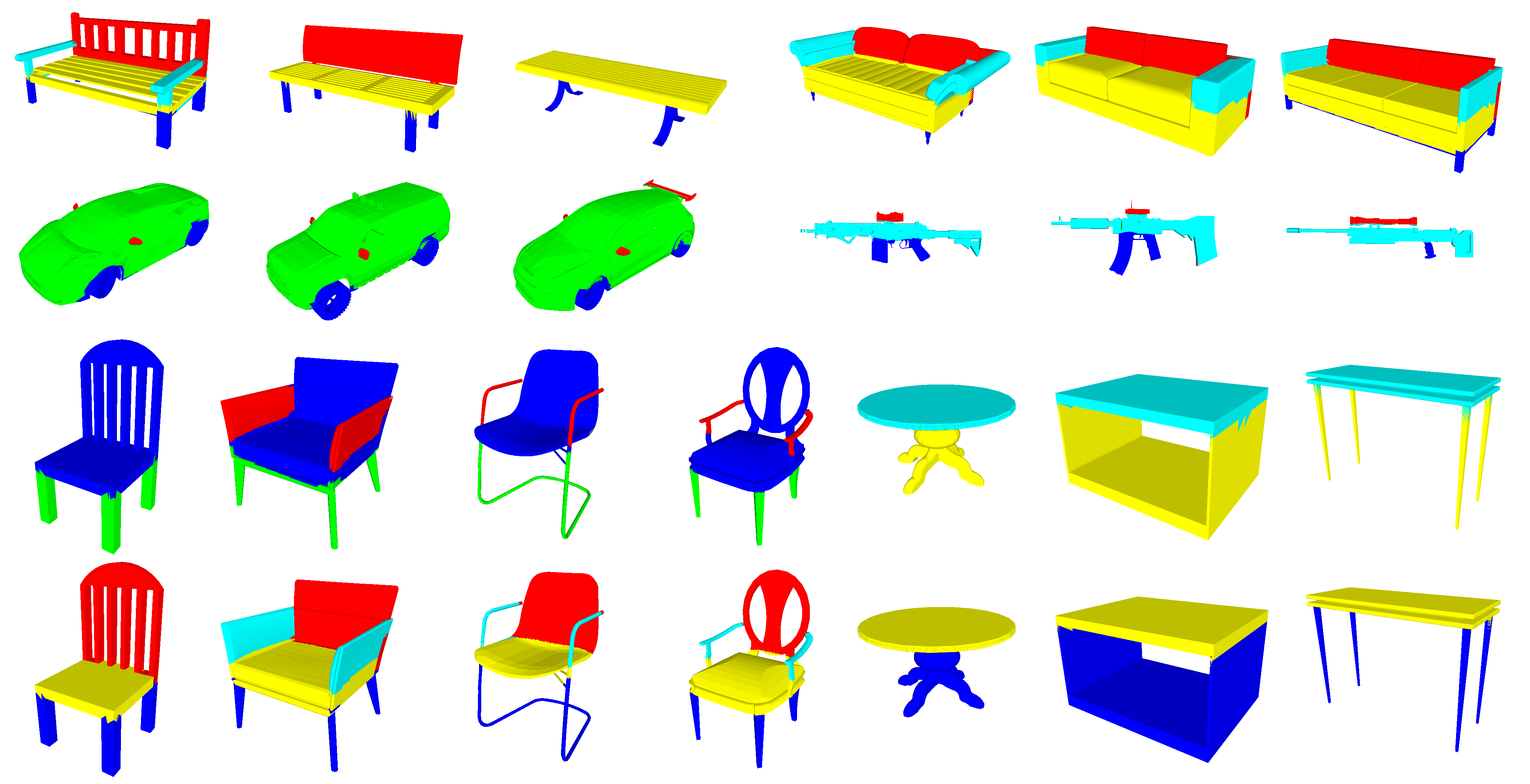}
\end{center}
   \caption{Unsupervised segmentation results by \baenet. The first three rows show segmentation results on bench, couch, car, rifle, chair and table respectively. In the last row, we show the results when merging chair and table into a joint dataset and training on it. Since our model generates a field for each part, we render the original meshes with different colors representing different parts.}
\label{fig:unsup_result}
\vspace{-2mm}
\end{figure}

\begin{table*}[t!]
\begin{center}
\begin{tabular}{l|c|c|c|c|c|c|c|c}
\hline
Shape (\#parts)  & airplane (3) & bag (2) & cap (2) & chair (3) & chair* (4) & mug (2) & skateboard (2) & table (2) \\
\hline
Segmented & {\small body, tail,} & {\small body,} & {\small panel,} & {\small back+seat,} & {\small back, seat,} & {\small body,} & {\small deck,} & {\small top,} \\
parts & {\small wing+engine} & {\small handle} & {\small peak} & {\small leg, arm} & {\small leg, arm} & {\small handle} & {\small wheel+bar} & {\small leg+support} \\
\hline
IOU & 61.1 & {\bf 82.5} & {\bf 87.3} & 65.5 & {\bf 83.7} & {\bf 93.4} & 63.5 & 78.7 \\
\hline
mod-IOU & {\bf 80.4} & {\bf 82.5} & {\bf 87.3} & {\bf 86.6} & {\bf 83.7} & {\bf 93.4} & {\bf 88.1} & {\bf 87.0} \\
\hline
\end{tabular}
\end{center}
\caption{ Quantitative results by \baenet on the ShapeNet part dataset~\cite{yi2016scalable} by IOU meansured against ground-truth parts. Chair* is chair training on chair+table joint set. mod-IOU, or modified IOU, is IOU measured against both parts and part combinations in the ground truth; it is more tolerant with coarse segmentations, e.g., combining the back and seat of a chair. Higher IOU indicates better performance.}
\label{table:unsup_number}
\vspace{-3mm}
\end{table*}

We first test unsupervised co-segmentation over 20 categories, where 16 of them are from the ShapeNet part dataset ~\cite{yi2016scalable}. These categories and the number of shapes are: planes (2,690), bags (76), caps (55), cars (898), chairs (3,758), earphones (69), guitars (787), knives (392), lamps (1,547), laptops (451), motors (202), mugs (184), pistols (283), rockets (66), skateboards (152), and tables (5,271). The 4 extra categories, benches (1,816), rifles (2,373), couches (3,173), and vessels (1,939), are from ShapeNet~\cite{chang2015shapenet}. We train individual models for different shape categories. 

Figures~\ref{fig:teaser} and~\ref{fig:unsup_result} show some visual results, with more in the supplemental material. Reasonable parts are obtained, and each branch of \baenet only outputs a specific part \rz{with a designated color}, giving us natural part correspondence. Our unsupervised segmentation is not guaranteed to produce the same part counts as those in the ground truth; it tends to produce coarser segmentations, e.g., combining the seat and back of a chair. Since coarser segmentations are not necessarily wrong results, in Table~\ref{table:unsup_number}, we report two sets of Intersection over Union (IOU) numbers which compare segmentation results by \baenet and the ground truth, one allowing part combinations and the other not.


Although unsupervised \baenet may not separate chair backs and seats when trained on chairs, it can do so when tables are added for training. Also, it successfully corresponds chair seats with table tops, chair legs with table legs; see Figure~\ref{fig:unsup_result}. This leads to a weakly supervised way of segmenting target parts, as we discuss next.

\subsection{Comparison with Tags2Parts}
\label{subsec:weaklysupervised_segmentation}

\begin{figure}[t!]
\begin{center}
\includegraphics[width=1.0\linewidth]{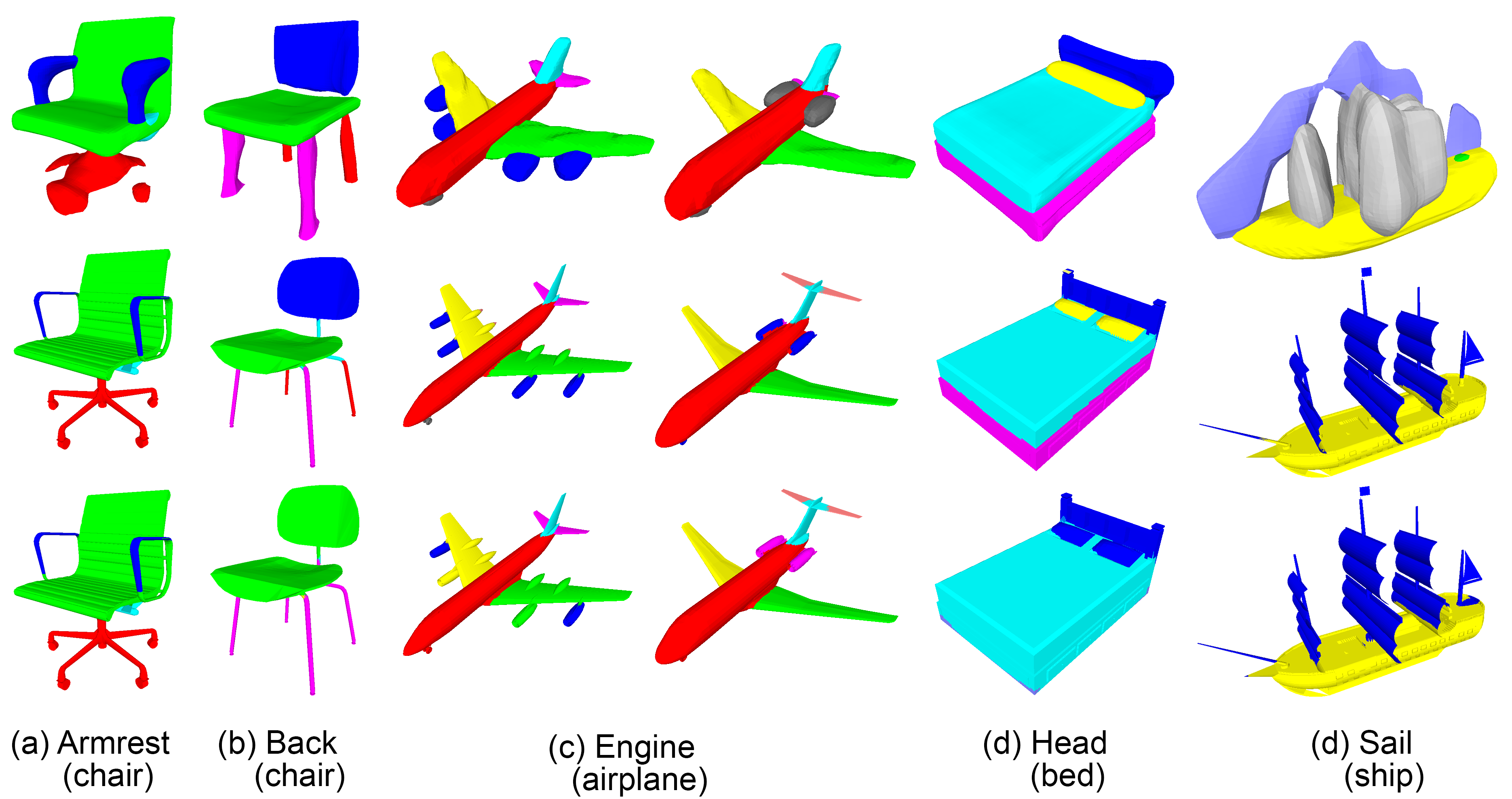}
\end{center}
   \caption{Weakly supervised segmentation results on Tags2Parts datasets~\cite{tags2parts}. Top row visualizes the implicit field of each branch by its 0.4-isosurface; different colors reflect outputs from different branches. This visualization \rz{is imperfect\/} 
since the field is not necessarily zero in empty areas. Middle row shows actual labelings assigned by the implicit fields: target parts in blue. Bottom row shows results of unsupervised training, i.e., without changing the shape distribution of the given dataset by per-shape labels. Some parts are not separated compared to weakly supervised results.}
\label{fig:tags2parts_result}
\vspace{-2mm}
\end{figure}

\begin{table}[t!]
\begin{center}
\begin{tabular}{l|c|c|c|c|c}
\hline
  & Arm & Back & Engine & Sail & Head \\
\hline\hline
Tags2Parts~\cite{tags2parts} & 0.71 & 0.79 & 0.46 & 0.84 & 0.37 \\
\baenet & {\bf 0.94} & {\bf 0.85} & {\bf 0.88} & {\bf 0.92} & {\bf 0.76} \\
\hline
\end{tabular}
\end{center}
\caption{Comparison with Tags2Parts~\cite{tags2parts} on their datasets by AUC (higher number = better performance). \baenet outperforms~\cite{tags2parts} in every category, even though our network did not use the provided per-shape labels explicitly.}
\label{table:tags2parts_number}
\vspace{-3mm}
\end{table}

We compare \baenet with the state-of-the-art weakly supervised part labeling network, Tags2Parts~\cite{tags2parts}. Given a shape dataset and a binary label for each shape indicating whether a target part appears in it or not, Tags2Parts can separate out the target parts, with the binary labels as weak supervision. \baenet can accomplish the same task with even weaker supervision. We do not pass the labels to the network or incorporate them into the loss function. Instead, we use the labels to change the training data distribution.

Our intuition is that, if two parts always appear together and combine in the same way, like chair back and seat, treating them as one single part is more efficient for the network to reconstruct them. But when we change the data distribution, e.g., letting only 20\% of the shapes have the target part (such as chair backs), it will be more natural for the network to separate the two parts and reconstruct the target part in a single branch. Therefore, we add weak supervision by simply making the number of shapes that do not have the target part four times as many as the shapes that have the target part, by duplicating the shapes in the dataset.

We used the dataset provided by~\cite{tags2parts}, which contains six part categories: (chair) armrest, (chair) back, (car) roof, (airplane) engine, (ship) sail, and (bed) head. We run our method on all categories except for car roof, since it is a flat surface part that our network cannot separate. We used the unsupervised version of our network to perform the task, first training for a few epochs using the distribution altered dataset for initialization, then only training our network on those shapes that have target parts to refine the results.

To compare results, we used the same metric as in~\cite{tags2parts}: Area Under the Curve (AUC) of precision/recall curves. For each test point, we get its probability of being in each part by normalizing the branch outputs with a unit sum. Quantitative and visual results are shown in Table~\ref{table:tags2parts_number} and Figure~\ref{fig:tags2parts_result}. Note that some parts, e.g., plane engines, that cannot be separated when training on the original dataset are segmented when training on the altered dataset. Our network particularly excels at segmenting planes, converging to eight effective branches representing body, left wing, right wing, engine and wheel on wings, jet engine and front wheel, vertical stabilizer, and two types of horizontal stabilizers.

\begin{figure}[t!]
\begin{center}
\includegraphics[width=1.0\linewidth]{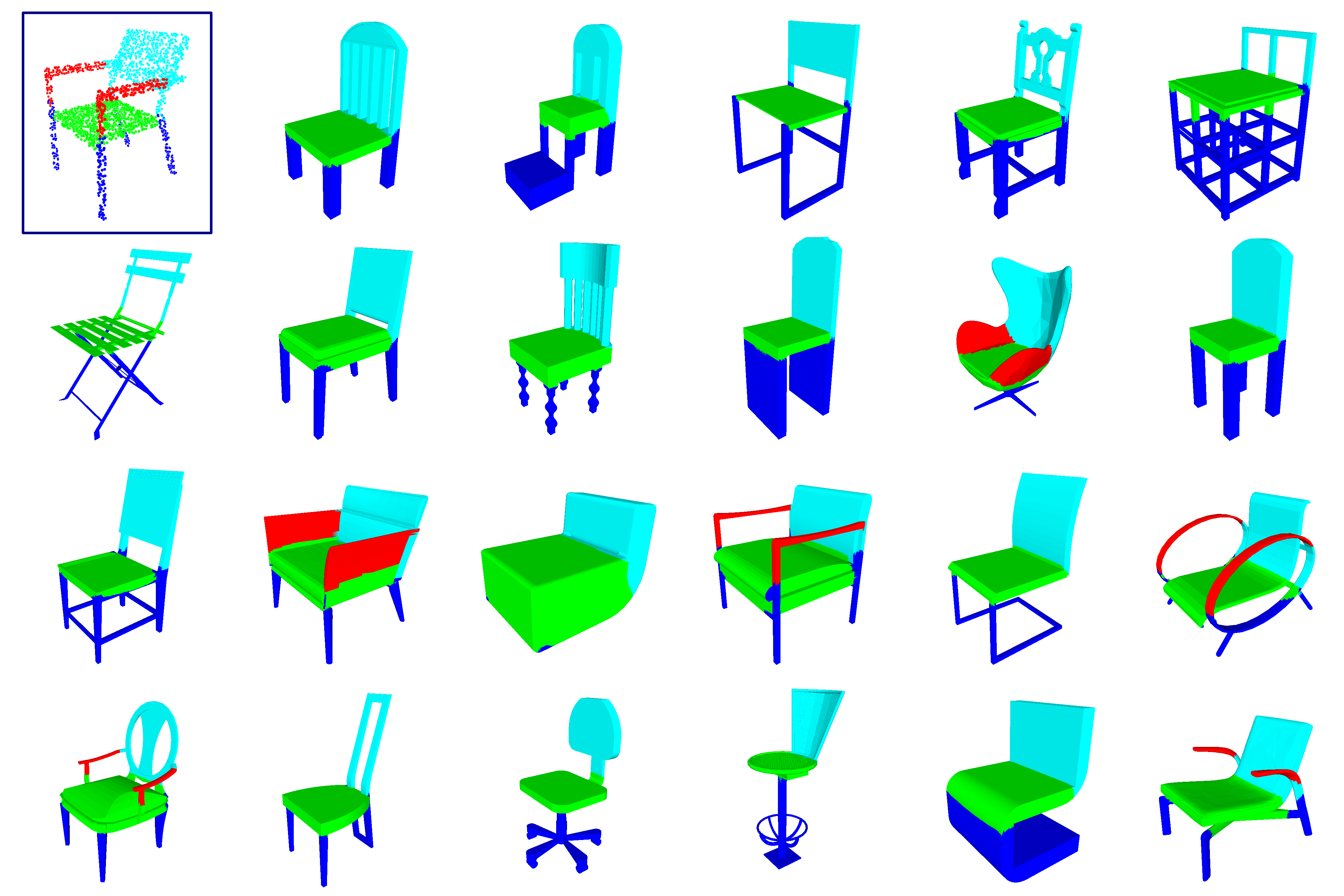}
\end{center}
\setlength{\belowcaptionskip}{-8pt}
   \caption{One-shot segmentation results by \baenet, with one segmented exemplar (blue box). See examples of other categories and 2/3-shot training results in the supplementary material.}
\label{fig:oneshot_result}
\vspace{-5mm}
\end{figure}

\begin{table}[t!]
\begin{center}
\setlength{\tabcolsep}{0.25em}
\begin{tabular}{l|c|c|c}
\hline
  & 1-exem. vs. & 2-exem. vs. & 3-exem. vs. \\
  & {\small 10\%} train set & {\small 20\%} train set & {\small 30\%} train set \\
\hline\hline
Pointnet~\cite{pointnet} & 72.1 & 73.0 & 74.6 \\
Pointnet++~\cite{pointnet++} & 73.5 & 75.4 & 76.6 \\
PointCNN~\cite{pointcnn} & 58.0 & 65.6 & 65.7 \\
SSCN~\cite{SSCN} & 56.7 & 61.0 & 64.6 \\
Our \baenet & {\bf 76.6} & {\bf 77.6} & {\bf 78.7} \\
\hline
\end{tabular}
\end{center}
\caption{Quantitative comparison to supervised methods by average IOU over 15 shape categories, without combining parts in the ground truth. Our one-shot learning with 1/2/3 exemplars outperforms supervised methods trained on 10\%/20\%/30\% of the shapes, respectively \rz{(on average each category has 765 training shapes).}}
\label{table:oneshot_number}
\vspace{-3mm}
\end{table}

\begin{figure}[t!]
\begin{center}
\includegraphics[width=1.0\linewidth]{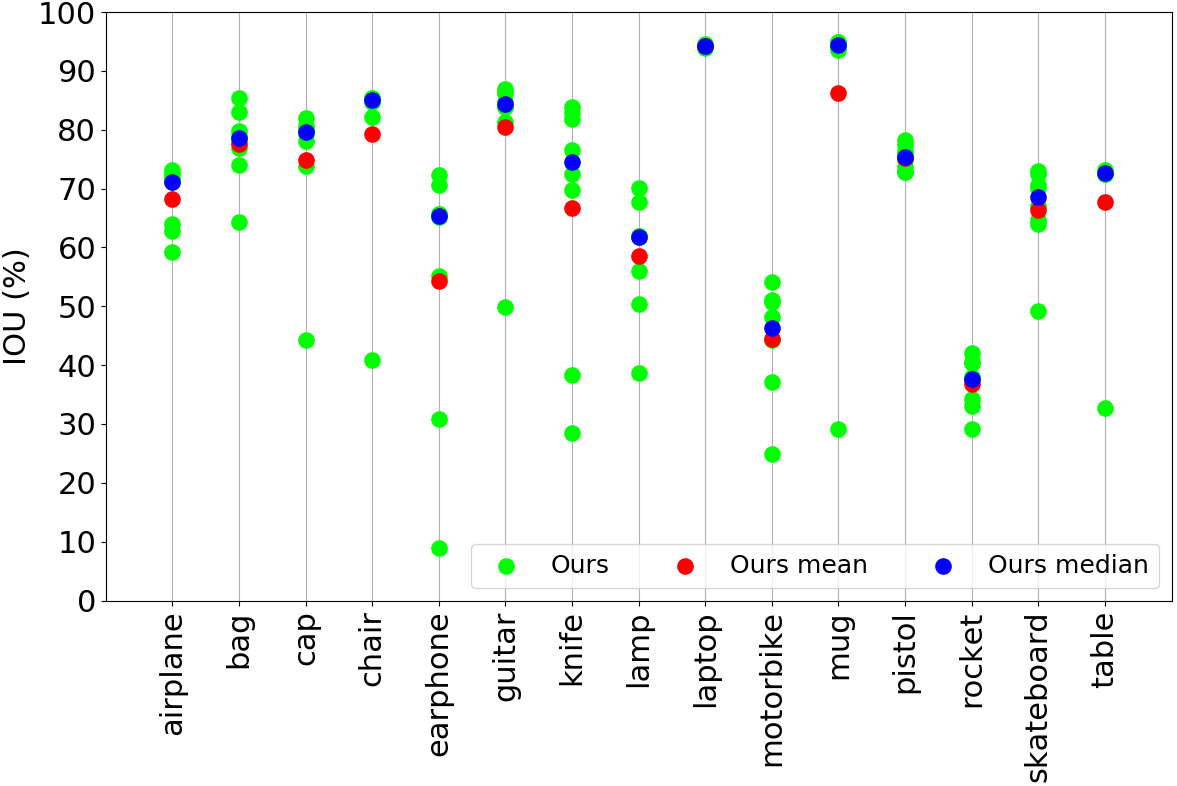}
\end{center}
\setlength{\belowcaptionskip}{-8pt}
\vspace{-10pt}
   \caption{\rz{Per-category, 1-exemplar results by \baenet with 8 randomly selected exemplars. Each dot plots an average IOU.}}
\label{fig:oneexample_number}
\vspace{-5mm}
\end{figure}

\subsection{One-shot training vs.~supervised methods}
\label{subsec:oneshot_segmentation}

\rz{Finally, we select few, e.g., 1-3, segmented exemplars from the training set to enforce \baenet to output designated parts. This allows us to evaluate \baenet using ground truth labels and compare to other methods. We train our model on the exemplar shapes using the supervised loss, while training on the whole set using unsupervised loss.}

\rz{Our evaluation is mainly against supervised methods since there has been hardly any semi-supervised segmentation methods that take only a few exemplars and segment 
shapes in a whole set. An exception is the very recent work by Zhao et al.~\cite{zhao20193d} which used only 5\% of the training data. In comparison, their IOU is 70\% averaged over the 
shapes, while our 1-exemplar result is 73.5\% even by setting all IOUs of cars to zero. Next, we compare \baenet to several state-of-the-art supervised methods including PointNet~\cite{pointnet}, PointNet++~\cite{pointnet++}, PointCNN~\cite{pointcnn}, and SSCN~\cite{SSCN}.}
Since it would be unfair to provide the supervised methods with only 1-3 exemplars to train, as we did for \baenet, we train their models using 10\%, 20\%, or 30\% of the shapes from the datasets for comparison, \rz{where, on average, there are 765 training shapes per category.} We evaluate all methods on the ShapeNet part dataset~\cite{yi2016scalable} by average IOU (no part combinations are tolerated), and train individual models for different shape categories. We did not include the car category for the same reason as in Section~\ref{subsec:weaklysupervised_segmentation}.

\rz{Table~\ref{table:oneshot_number} shows quantitative comparison results, averaged over 15 shape categories, and some visual results for the chair category are provided in Figure~\ref{fig:oneshot_result}. As we can see, \baenet trained with 1/2/3 exemplars outperforms supervised methods with 10\%/20\%/30\% of the training set, respectively. 
Note that we trained the supervised models with their original codes and parameters, which may be a better fit for larger training sets. Hence, we performed additional experiments (1-exemplar vs.~10\% train data) by reducing the network capacity (to $\frac{1}{8}$, $\frac{2}{8}$, ... ,$\frac{7}{8}$) and adding regularization (with scales $10^{-3}$, $10^{-4}$, ... , $10^{-10}$). The best results obtained were: PointNet 73.8\%, PointNet++ 74.0\%, PointCNN 58.6\%, and SSCN 57.8\%, which are still no better than \baenet. Per-category comparison results can be found in the supplementary material, where we also show results of changing the number of network layers.}

\rz{In general, the performance of one- or few-shot learning hinges on few exemplars, hence their selection does make a difference. In Figure~\ref{fig:oneexample_number}, we show all the 1-exemplar results by using 8 randomly selected exemplars, while Table~\ref{table:oneshot_number} took the best performing exemplar for each category. As we can see, result sensitivity depends on the shape category. For some categories, the numbers are consistent (e.g., around 94\% for laptops), while for others, the numbers can vary a lot (e.g., 8.9\% to 72.3\% for earphones), owing to shape variations in the category. For earphones, the outlier is due to an exemplar which has a long cord that makes the shape smaller and misaligned after normalization.}

Ideally, we hope to select exemplars that contain all the ground truth parts and are representative of the shape collection, but this can be difficult in practice. For example, there are two kinds of lamps -- ground lamps and ceiling lamps -- that can both be segmented into three parts. However, the order and labels of the lamp parts are different, e.g., the base of a ground lamp and the base of a ceiling lamp have different labels. Our network cannot infer such semantic information from 1-3 exemplars, thus we select only one type of lamps (ground lamps) as exemplars, and our network only has three branches (without one for the base of the ceiling lamp). During evaluation, we add a fake branch that only outputs zero, to represent the missing branch.


\section{Conclusion, limitations, and future work}
\label{sec:future}

We have introduced \baenet, a branched autoencoder, for unsupervised, one-shot, and weakly supervised shape co-segmentation. Experiments show that our network can outperform state-of-the-art supervised methods, including PointNet++, PointCNN, etc., using much less training data (1-3 exemplars vs.~77-230 for the supervised methods, average over 15 shape categories). On the other hand, compared to the supervised methods, \baenet tends to produce coarser segmentations, which are correct and can provide a good starting point for further refinement.

Many prior unsupervised co-segmentation methods~\cite{golovinskiy2009consistent,hu2012co,huang2011joint,sidi2011unsupervised}, which are model-driven rather than data-driven, had only been tested on very small input sets (less than 50 shapes). In contrast, \baenet can process much larger collections (up to 5,000+ shapes). In addition, unless otherwise noted, all the results shown in the paper were obtained using the default network settings, further validating the generality and robustness of our co-segmentation network.

\baenet is able to produce consistent segmentations, over large shape collections, without explicitly enforcing a consistency loss. In fact, the consistency is a consequence of the network architecture. However, our current method does not provide any theoretical guarantee for segmentation consistency or universal part counts; such extensions could build upon the results of \baenet. \rz{Similar
to prior works on co-segmentation, we assume that shapes in the input collection are consistently aligned. Learning rotation-invariant deep models for the task is interesting future work.}


For unsupervised segmentation, since we initialize the network parameters randomly and optimize a reconstruction loss, while treating each branch equally, there is no easy way to predict which branch will output which part.
%
%
The network may also be sensitive to the initial parameters, where different runs may result in different segmentation results, e.g., combining seat and back vs.~seat and legs for the chair category. Note however that both results may be acceptable as a coarse structural interpretation for chairs.


Another drawback is that our network groups similar and close-by parts in different shapes for correspondence. This is reasonable in most cases, but for some categories, e.g., lamps or tables, where the similar and close-by parts may be assigned different labels, our network can be confused. How to incorporate shape semantics into \baenet is worth investigating.
Finally, \baenet is much shallower and thinner compared to IM-NET~\cite{IMGAN}, since we care more about segmentation (not reconstruction) quality. However, the limited depth and width of the network make it difficult to train on high-resolution models (say $64^3$), which hinders us from obtaining fine-grained segmentations.


In future work, besides addressing the issues above, we plan to introduce hierarchies into the shape representation and network structure, since it is more natural to segment shapes in a coarse-to-fine manner. Also, \baenet provides basic part separation and correspondence, which could be incorporated when developing generative models.

\vspace{5pt}
\noindent
\rz{{\bf Acknowledgment.}~We thank the anonymous reviewers for their valuable feedback, Vova Kim and Daniel Cohen-Or for discussions, and Wallace Lira for proofreading. This research is supported by NSERC and an Adobe gift.}

{\small
\bibliographystyle{ieee_fullname}
\bibliography{coseg}
}

\end{document}